\documentclass[conference]{IEEEtran}
\IEEEoverridecommandlockouts

\usepackage[hidelinks]{hyperref} % hidelinks to remove url bounding box
\usepackage{orcidlink}
\usepackage{amsmath,amssymb,amsfonts}
\usepackage{enumitem}
\setlist{nosep}
\usepackage{url}
\usepackage{algorithm}
\usepackage{algpseudocode}
\usepackage{booktabs}
\usepackage{authblk}
\usepackage{balance}
\begin{document}

\title{GenFlow: Interactive Modular System for Image Generation}

\author[1, 2]{Duc-Hung Nguyen\orcidlink{0009-0008-9944-1581}$^\dagger$\thanks{$\dagger$Both authors contributed equally to this research.}}
\author[1, 2]{Huu-Phuc Huynh\orcidlink{0009-0006-1170-7409}$^\dagger$}
\author[1, 2]{Minh-Triet Tran\orcidlink{0000-0003-3046-3041}}
\author[1, 2]{Trung-Nghia Le\orcidlink{0000-0002-7363-2610}$^\ddagger$\thanks{$\ddagger$Corresponding author. {\it e-mail: ltnghia@fit.hcmus.edu.vn}}}

\affil[1]{University of Science, Ho Chi Minh City, Vietnam}
\affil[2]{Vietnam National University, Ho Chi Minh City, Vietnam}

% \author{Anonymous Authors}

% \makeatletter
% \g@addto@macro\@maketitle{
% % \vspace{-13mm}
% % \centering \textbf{\url{https://selab.hcmus.edu.vn/}}
% % \vspace{-3mm}
%   \begin{figure}[H]
%   \setlength{\linewidth}{\textwidth}
%   \setlength{\hsize}{\textwidth}
%   \centering
% %   \rule{10cm}{5cm} % this is your image
%   \includegraphics[width=1\linewidth]{figures/teaser_3.png}
%     \vspace{-3mm}
%     \caption{GenFlow interface, including two key components: Canvas for efficient workflow creation and management and Flow Pilot for smart workflow search.}
%     % \Description{The GenFlow interface overview includes the following: on the left, there is the Canvas, while on the right, the sidebar contains Flow Pilot, a smart assistant. The remaining tabs in the sidebar are used for controlling and editing workflows in the Canvas.}
% 	\label{fig_teaser}
%   \end{figure}
% }
% \makeatother

\maketitle

\begin{abstract}
Generative art unlocks boundless creative possibilities, yet its full potential remains untapped due to the technical expertise required for advanced architectural concepts and computational workflows. To bridge this gap, we present GenFlow, a novel modular framework that empowers users of all skill levels to generate images with precision and ease. Featuring a node-based editor for seamless customization and an intelligent assistant powered by natural language processing, GenFlow transforms the complexity of workflow creation into an intuitive and accessible experience. By automating deployment processes and minimizing technical barriers, our framework makes cutting-edge generative art tools available to everyone. A user study demonstrated GenFlow’s ability to optimize workflows, reduce task completion times, and enhance user understanding through its intuitive interface and adaptive features. These results position GenFlow as a groundbreaking solution that redefines accessibility and efficiency in the realm of generative art.
  
\end{abstract}


\begin{figure*}[t!]
    \centering
    \includegraphics[width=1\textwidth]{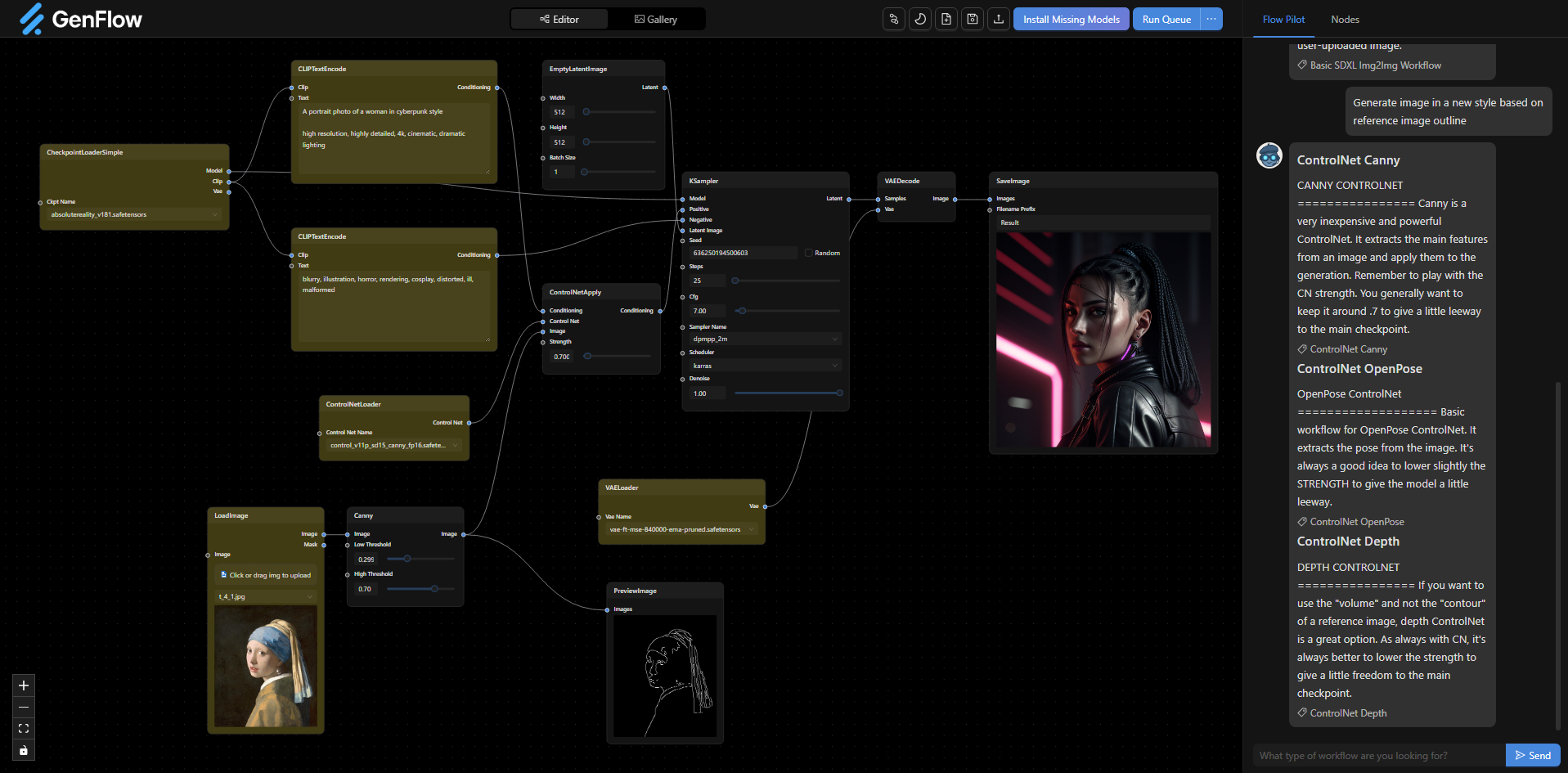}
    \vspace{-3mm}
    \caption{GenFlow interface, including two key components: Canvas for efficient workflow creation and management and Flow Pilot for smart workflow search.}
    \label{fig_teaser}
\end{figure*}

\section{Introduction}
Generative art has experienced a surge in popularity, driven by advances in machine learning and vision-based models. From editing and advertising to the film industry, AI-powered image-generation tools are transforming creative workflows \cite{chatterjee2022art}. Popular platforms like Stable Diffusion WebUI, Midjourney, and DALL-E 3 offer powerful capabilities. However, their limited interfaces can pose challenges for tasks that require greater customization, such as style transfer, image composition, or applying advanced filters and effects \cite{10204579}.

%While tools like ComfyUI provide enhanced flexibility through graphical interfaces for creating and manipulating models, they often come with a steep learning curve, especially for users unfamiliar with advanced techniques like IPAdapter or ControlNet. This complexity makes it difficult for beginners to create sophisticated workflows without significant prior knowledge.
Tools such as ComfyUI\footnote{\url{https://www.comfy.org/}} have emerged to address this, offering enhanced flexibility through node-based visual programming. In this visual programming, source code is presented in a graph-based format, called \textit{workflow}, where each node (i.e., the unit of code that performs a specific task) is a vertex and each connection (i.e., the flow of data from one node to another) is an edge. Although these tools provide robust customization options, they also introduce steep learning curves, which require significant machine learning techniques like IPAdapter~\cite{ye2023ip} or ControlNet~\cite{zhang2023adding}. As a result, many users, particularly novices or those without a technical background, cannot fully leverage the potential of these advanced tools, leaving a gap in accessibility and usability \cite{10.1145/3638884.3638908}.

To bridge this gap, we introduce GenFlow, a novel modular framework designed to make advanced image generation tools more accessible while maintaining a high level of customization \cite{wingstrom2024redefining}. GenFlow integrates two core components: Editor is a user-friendly node-based interface to create and manage workflows, catering to users with varying levels of expertise; Flow Pilot is an intelligent assistant powered by natural language processing that simplifies workflow discovery and deployment by allowing users to describe their desired tasks in natural language \cite{ramesh2021zero}. In addition, GenFlow incorporates features such as automated model retrieval, a local database for faster workflow search, and intelligent web exploration to enhance efficiency and usability. By removing technical barriers, the system empowers users to easily explore generative art tools, fostering creativity and experimentation \cite{9432822}.

%To simplify this process and make advanced image generation more accessible, we introduce \textit{GenFlow}, an Interactive Modular Framework for Image Generation. GenFlow consists of two main components: \textit{Editor} is a node-based interface designed for intuitive workflow customization, catering to users of all skill levels, and \textit{Flow Pilot} is an integrated chatbot that uses natural language processing to assist users in finding and applying suitable workflows efficiently. GenFlow also features automated model search and download capabilities, reducing reliance on manual browsing. By incorporating a local database to store temporary workflows, the system accelerates the search process and ensures improved semantic matching, significantly enhancing usability.

We evaluate the effectiveness of GenFlow through a user study, and highlight its potential to redefine accessibility in generative art. In this user study, participants praised GenFlow for streamlining the workflow selection process and offering an accessible yet powerful tool for image generation. %Their feedback highlighted both the benefits and areas for improvement, informing future development.

Our contributions are as follows:
\begin{itemize}
    \item We introduce GenFlow, a user-friendly system for customizable image generation, suitable for both beginners and experts.
    \item We develop Flow Pilot, an intelligent assistant for natural language-based workflow discovery.
    \item A user study demonstrates the effectiveness and usability of the proposed system.
\end{itemize}

\section{Related Work}

\subsection{Node-Based Workflow Systems}

Node-based visual programming environments have proven effective in various domains, including 3D modeling (Blender's node editor %\footnote{https://www.blender.org}
), game development (Unreal Engine's Blueprints %\footnote{https://www.unrealengine.com/fr/blog/introduction-to-blueprints}
), and audio processing (Max/MSP %\footnote{https://cycling74.com}
) \cite{angert2023spellburst} \cite{bieniek2024generativeaimultimodaluser}. These systems allow users to create complex processes by connecting nodes representing different operations. This approach offers a high degree of flexibility and control but can also present a steep learning curve for new users \cite{noone2018visual}. ComfyUI %\footnote{https://www.comfy.org} 
 is a notable example of a node-based interface for Stable Diffusion \cite{DBLP:journals/corr/abs-2112-10752}. While offering powerful customization options, it requires significant technical knowledge of Stable Diffusion pipelines, making it less accessible to casual users.

\subsection{Retrieval Augmented Generation (RAG)}

RAG combines the strengths of information retrieval and generative models. It improves the quality and factual accuracy of generated text by grounding it in external knowledge sources \cite{lewis2021retrievalaugmentedgenerationknowledgeintensivenlp} \cite{gao2023retrieval}. RAG has been successfully applied in various NLP tasks, such as question-answering~\cite{he2024g} and dialogue generation~\cite{wang2024unims}. In our work, we adapt the RAG paradigm to the domain of AI art generation by retrieving relevant workflows based on user-provided descriptions, bridging the gap between natural language user input and structured workflow execution \cite{Hameed01012021}.

\subsection{Web Search}

The field of autonomous web agents has seen remarkable progress with the advent of large language models (LLMs), which enable the automation of complex tasks on digital platforms, ranging from web navigation to interaction with graphical user interfaces (GUIs).  For example, multimodal agents such as WebVoyager leverage combined vision–language understanding to identify interactive elements on a page, although they sometimes fall short of grasping the semantic intent behind those elements \cite{he2024webvoyager}.  Building on this, systems like OmniParser go further by parsing UI components to infer their functions, yet they may still overlook certain interactable widgets in dynamic or deeply nested layouts \cite{lu2024omniparser}.  Complementing these modeling advances, benchmark suites such as VisualWebArena have been introduced to evaluate agent performance in realistic web scenarios, and efforts like SeeClick target improvements in element detection and grounding to ensure more reliable action selection during browsing tasks \cite{koh-etal-2024-visualwebarena, cheng-etal-2024-seeclick}.  Together, these contributions lay the groundwork for the Web Suffer Agent, which seeks to integrate multimodal recognition, semantic UI parsing, and robust benchmarking into a unified framework for more effective and generalizable web automation.

GenFlow extends prior node-based systems by introducing a natural language interface that lowers the barrier for non-expert users while retaining the flexibility required by advanced creators. Unlike existing tools that focus solely on visual programming or textual interaction, GenFlow combines both through Flow Pilot, enabling intuitive workflow discovery and customization. Additionally, the multi-agent web exploration engine enhances existing web automation efforts by tailoring retrieval specifically for image generation tasks. This unified approach makes GenFlow a more accessible and intelligent solution for generative image creation.

\begin{figure*}[t!]
\centering
\includegraphics[angle=90, width=\textwidth]{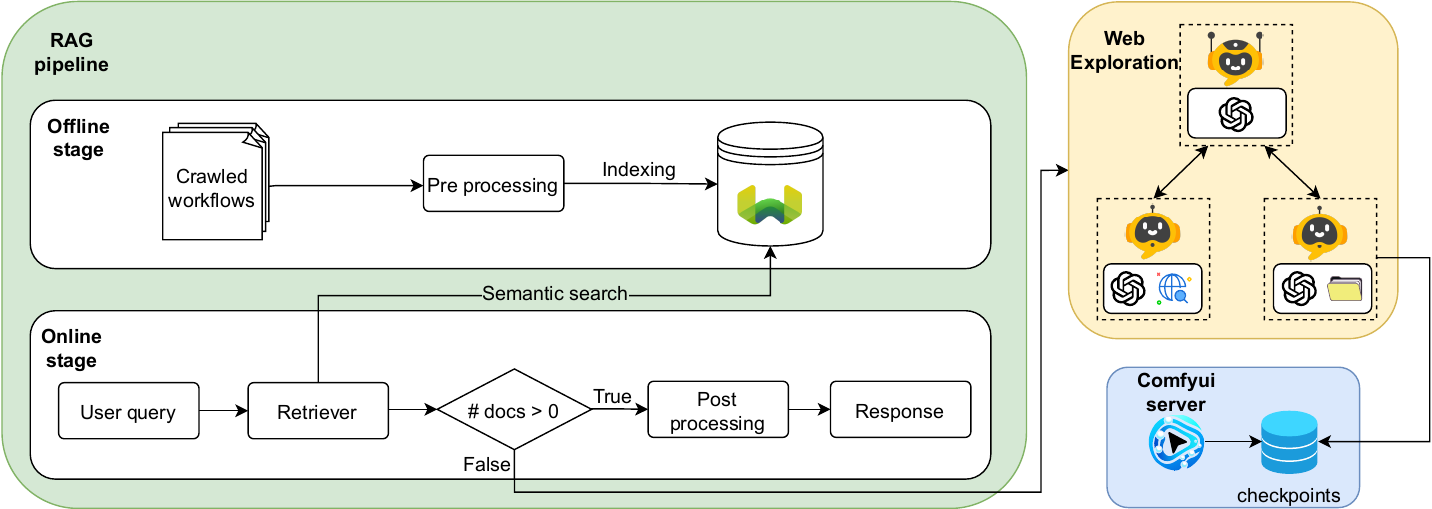}
% \caption{The framework operates through two distinct phases: \textbf{(a) Offline stage} -- workflows are crawled, preprocessed, and indexed to construct a vector database for semantic retrieval; \textbf{(b) Online stage} -- user queries are processed via a retriever module that performs semantic search against the indexed knowledge base. If relevant documents are found, the system generates responses; otherwise, it executes a web exploration module that automatically downloads workflow or checkpoints from the internet to the ComfyUI server for response generation.}
% \caption{Overview of the proposed system's workflow. \comment{The font size within the figure is also very small. Also, the caption should atleast be re-written to give more details into the workflow shown in the figure.}}
\caption{The framework operates through two distinct phases: \textbf{(a) Offline stage} - crawled workflows are preprocessed to remove noise (e.g., hashtags, links) and indexed into a Weaviate vector database. \textbf{(b) Online stage} - user queries trigger a semantic search; if relevant results are found, they are post-processed and returned. If no suitable workflows are retrieved, a hierarchical multi-agent web exploration system is activated. A supervisor agent coordinates two workers: one searches the web for relevant workflows and model checkpoints, while the other downloads model checkpoints and stores them in the correct folder structure required for ComfyUI to function properly.}
\label{fig:pilot_pipeline}
\vspace{-5mm}
\end{figure*}

\section{Proposed GenFlow System}

% \vspace{-20px}

% \subsection{Conceptualization}
% \label{sec:conceptualization}

% In light of the aforementioned challenges of ...., we introduce .... aimed at .... We conceptualize ... as an application designed to aid users in easily creating their ....

% In light of the aforementioned challenges of lack fine-grained control for expert users, while node-based systems like ComfyUI \cite{ComfyUI2023} have a steep learning curve for end user, we introduce GenFlow, an novel interactive modular framework for image generation, aimed at making workflow creation and selection more accessible. We conceptualize this framework as an application designed to aid users in easily creating their desired image generation pipelines using a node/graph-based interface, with the added benefit of chatbot-assisted workflow retrieval and automated web-based workflow search.

\subsection{User Interface}

GenFlow is an interactive modular framework that streamlines the creation and discovery of image generation workflows. It merges a node-based visual editor with a natural language assistant, allowing users of all skill levels to build complex pipelines with minimal technical effort. Unlike existing tools that prioritize flexibility at the expense of usability, GenFlow lowers the barrier to entry while preserving advanced customization. These goals are realized through a unified interface that integrates visual programming with intelligent workflow retrieval. The interface consists of two main components, as shown in Figure \ref{fig_teaser}:

\textit{Canvas}: The canvas provides a visual programming space where users can drag and drop nodes to construct workflows. Each node represents a task (e.g., text-to-image, image enhancement), and users can define parameters, upload media, and establish connections. A sidebar includes a “Nodes” tab to browse available operations and a “Gallery” tab to save and manage generated outputs and workflows. This setup supports fast iteration and encourages exploration through a clear, visual structure.

% Canvas contains current workflow. A sidebar provides two tabs: "Node" and "Workflow," which allow users to access a node library and manage workflows. The canvas allows users to drag and drop nodes to create or edit workflows. Users can create and modify connections by dragging from the target handle of one node to the source handle of another. Additionally, users can upload images, edit text, and configure parameters for individual nodes. Next to the "Editor" tab is the "Gallery," where all generated images are saved.

% On the sidebar, users can click on the "Nodes" tab to view the installed nodes in the system. From there, they can drag a node onto the canvas to deploy it. Next to the "Nodes" tab is the "Workflows" tab, where users can save, download, upload, and manage their workflows.

\textit{Flow Pilot}: This assistant enables users to describe tasks in plain language, such as “generate a portrait using this sketch.” The system semantically matches the input with relevant workflows, ranks them by relevance and popularity, and displays descriptions for review. Users can deploy a selected workflow directly onto the canvas, greatly accelerating the process of discovering and customizing image generation tasks.

% \begin{figure*}[ht]
% \centering
% \begin{subfigure}[b]{0.48\textwidth}
%     \centering
%     \includegraphics[width=\textwidth]{figures/ui_node-picker.png}
%     \caption{Node picker sidebar}
%     \label{fig:ui_node-picker}
% \end{subfigure}
% \hfill % Important for spacing between subfigures
% \begin{subfigure}[b]{0.48\textwidth} % Adjust width as needed
%     \centering
%     \includegraphics[width=\textwidth]{figures/teaser.png} % Replace with your second image
%     \caption{Overview}
%     \label{fig:overview}
% \end{subfigure}
% \caption{User Interface Components} % Overall caption for the figure
% \label{fig:ui_components} % Overall label for the figure
% \end{figure*}

\subsection{System Workflow}

\subsubsection{Overview}

The core of our system is powered by the Stable Diffusion engine, which handles the execution of image generation pipelines. We build on the ComfyUI backend to manage these pipelines, extending its capabilities with automated support for custom node integration via the ComfyUI Manager. As illustrated in Fig.~\ref{fig:pilot_pipeline}, the ComfyUI Manager verifies and installs required models and nodes by retrieving installation metadata (e.g., URLs and save paths) from a local database. If the necessary assets are not found, control is handed back to ComfyUI, which then invokes the Web Exploration module to search for and download the missing components. Once retrieved, these resources are stored in the database and made immediately available within the ComfyUI environment.

\subsubsection{RAG Pipeline}

To ensure high-quality input for embedding and retrieval, we preprocess the description field by removing noise such as URLs, HTML tags, email addresses, social media references, emojis, punctuation, stopwords, and special characters. This results in clean, semantically meaningful text suitable for downstream analysis. Embeddings are generated using a configurable model, with GoogleAI's "models/embedding-001" as the default, and are stored in Weaviate, a vector database optimized for semantic search and metadata filtering.

User queries are processed in the same way and matched against stored entries using a similarity threshold to ensure relevance. Retrieved workflows are then ranked based on popularity indicators, such as the number of likes. A large language model (LLM) further refines and validates the top results, presenting users with curated workflow options that include names, descriptions, and metadata to support efficient discovery and deployment.

\subsubsection{Web Exploration Engine} 

Platforms such as WorkflowComfyUI, OpenArt, and Civitai have grown in popularity with the rise of generative AI, fostering communities where users share and explore workflows for image generation and editing. These platforms provide valuable repositories of creative and technical knowledge, enabling users to discover and adapt diverse ComfyUI workflows.

Despite their value, locating workflows for specific tasks across these platforms remains challenging. Users must often navigate different interfaces, understand platform-specific conventions, and manually review numerous options to identify suitable workflows. This process is time-consuming and introduces unnecessary friction to creative exploration. To overcome this barrier, we introduce a multi-agent web exploration engine (Fig. \ref{fig:pilot_pipeline}) that automates the discovery of task-relevant workflows based on user intent.

% Platforms such as WorkflowComfyUI, OpenArt, and Civitai have gained popularity alongside the rise of generative AI, fostering active communities where users share and explore workflows for tasks like image generation and editing. These platforms serve as valuable resources for discovering diverse ComfyUI workflows, enabling users to build on collective creativity and technical expertise.

% However, accessing workflows that suit specific tasks from various platforms is not straightforward. Users often need to understand the unique operations of each platform and invest significant time navigating and reviewing numerous options to find the desired workflow. This process is time-consuming and adds complexity to creativity and task optimization. To address this challenge, we propose using multi-agent as an intelligent web exploration engine (shown in Figure \ref{fig:pilot_pipeline}) to find workflows tailored to user requirements \cite{mazumder2020flin}.

% \begin{figure*}[ht]
% \centering
% \includegraphics[width=0.4\textwidth]{figures/websearch_pipeline.pdf}
% \caption{Web Exploration Pipeline}
% \label{fig:web_pipeline}
% \end{figure*}

\paragraph{Web Suffer Agent} 

\begin{algorithm}[t!]
\caption{Combine Interactive Elements}
\label{alg:combine_interactive_elements}
\begin{algorithmic}[1]
    \State \textbf{Input:} $W$ (WebVoyager elements), $O$ (OmniParser elements), $\tau$ (IoU threshold)
    \State \textbf{Output:} $C$ (combined elements list)
    \State \textbf{Initialize:} $C \gets [\,]$, $U \gets \emptyset$
    
    \ForAll{$w \in W$}
        \State $o^* \gets \arg\max_{o \in O \setminus U} \text{IoU}(w.\text{bbox}, o.\text{bbox})$
        \If{$o^*$ exists and $\text{IoU}(w.\text{bbox}, o^*.\text{bbox}) \geq \tau$}
            \State Add merged properties of $w$ and $o^*$ to $C$; $U \gets U \cup \{\text{index of } o^*\}$
        \Else
            \State Add $w$ to $C$
        \EndIf
    \EndFor

    \ForAll{$o \in O \setminus U$}
        \State Add $o$ to $C$
    \EndFor

    \State \Return $C$
\end{algorithmic}
\end{algorithm}

The Web Suffer Agent is designed to automatically gather relevant workflows from online platforms through web scraping, API calls, and optimized search queries. Traditional web agents often struggle with accurately identifying and interacting with visual elements due to limited perception and semantic understanding. To overcome these limitations, recent studies have proposed methods that enhance agents’ ability to parse and interpret complex web interfaces.

% WebVoyager \cite{he2024webvoyager} introduced a technique for extracting interactive elements from web pages using Set-of-Mark (SoM) prompting \cite{yang2023setofmark}, which improves element detection and interaction accuracy. While effective, this approach relies on heuristics and cannot infer element functionality or triggered events, forcing agents to depend on vision-based inference, which often leads to errors. To address this, OmniParser \cite{lu2024omniparser} was developed to infer UI element functionalities by learning from a diverse set of icon-action pairs. It improves semantic understanding but may still miss some interactable elements in complex layouts, limiting its coverage and reliability in dynamic interfaces.

WebVoyager \cite{he2024webvoyager} proposed using Set-of-Mark (SoM) prompting \cite{yang2023setofmark} to extract interactive elements from web pages, enhancing detection and interaction accuracy. However, this heuristic approach cannot infer element functionality, requiring agents to rely on error-prone vision-based inference. OmniParser \cite{lu2024omniparser} addresses this by learning element functionalities from icon-action pairs, but it may still miss some interactable elements in complex interfaces. By combining WebVoyager’s element detection with OmniParser’s functional inference, our agents gain both visual grounding and semantic context. This integration, illustrated in Algorithm~\ref{alg:combine_interactive_elements}, enables more accurate and comprehensive web interaction, marking a significant step toward robust and scalable web exploration in AI systems.

\paragraph{File Suffer Agent} 

The File Suffer Agent plays a key role in the multi-agent pipeline by retrieving, parsing, and organizing workflow data. It handles both existing files and those downloaded by the Web Suffer Agent, ensuring all relevant information is collected. The agent extracts workflow structures, parameters, and dependencies, converting unstructured data into a structured format for downstream analysis and integration.

% The File Suffer Agent plays a pivotal role in the multi-agent pipeline by managing the retrieval, parsing, and organization of workflow data. Its responsibilities include handling existing workflow files and those downloaded by the Web Suffer Agent to ensure all necessary data is collected and properly organized for subsequent processing. Additionally, it parses the retrieved files to extract essential information, such as workflow structures, parameters, and dependencies, transforming unstructured data into a structured format suitable for analysis and integration.
%     The File Suffer Agent is a crucial component in our multi-agent pipeline, responsible for managing the retrieval, parsing, and organization of workflow data. Its primary functions include:
% \begin{itemize}
%     \item \textbf{Data retrieval}: is responsible for managing existing workflow files as well as those downloaded by the Web Suffer Agent. It ensures that all required data is collected and organized properly for subsequent processing steps.
%     \item \textbf{Data parsing}: analyzes the retrieved files, extracting relevant information such as workflow structures, parameters, and dependencies.  This parsing process transforms unstructured data into a structured format suitable for analysis and integration.
    
% \end{itemize}

\paragraph{Multi-Agent System} 

Managing workflow discovery across diverse tasks is difficult for a single agent. To address this, we adopt a supervised multi-agent architecture where an Orchestration Agent delegates tasks to specialized worker agents, such as the Web Search Agent and File Suffer Agent \cite{delias2011agents}. Each agent maintains its own memory to track past actions and interactions, allowing the system to retain context, adapt to new tasks, and ensure accurate and efficient exploration. This design aligns with the Magentic-One framework \cite{fourney2024magentic-one}, which demonstrates the effectiveness of a lead orchestrator coordinating specialized agents to solve complex problems \cite{10.5555/1695886, li2012multi}.

\section{Comparison of Workflow Efficiency with Tradition Approach}
% \subsection{Comparision between our system and }
% Given task that user want to generate image base on another reference image and text

% Without our system: user have to search over the web or ask Chat AI how to process the task. Then following the technique and search related reposistory code. Then read document of the codebase and pull code from github or huggingface

% with out system: user dirrectly type the task to the FlowPilot, then it retrieve related workflow

% Overall, it tooks 12m47s for without system. And 3m13s with using our system

% To evaluate local database, we esiimate the time for FlowPilot retrieve workflow from user prompt with and without local database

% Both use same prompt "Give me basic workflows that used for convert image to image"

Without our system, users need to search the web or ask a chatbot for guidance to generate an image based on a reference image and conditioning text. This process involves finding suitable techniques, searching repositories, reviewing documentation, and downloading code and models from platforms like GitHub or Hugging Face. On average, this takes about 12 minutes and 47 seconds.

In contrast, with our system, users can directly input the task into FlowPilot, which instantly retrieves the relevant workflow. This streamlined approach reduces the time required to just 3 minutes and 13 seconds.

To evaluate the impact of using a local database, we measured retrieval times for the prompt: “Give me basic workflows used to convert image to image.” The time to install missing nodes and models varies depending on factors such as the number of missing components and their size. For example, as shown in Table~\ref{tab:local_database_comparison}, the “ThinkDiffusion—Img2Img” workflow retrieved via the web requires only one missing node and one model (ThinkDiffusionXL.safetensors with a size of 6.94 GB). This highlights how leveraging a local database significantly reduces the retrieval time (in milliseconds), offering users a faster and more efficient experience.

\begin{table}[t!]
\centering
\caption{Comparison of operations with and without a local database in term of milliseconds.}
% \vspace{-3mm}
\begin{tabular}{lcc}
\toprule
\textbf{} & \textbf{Local Database} & \textbf{Without Local Database} \\ \midrule
Search Workflow & 3,871 & 152,008 \\
Install Missing Nodes & 2,755 & 228,182 \\
Install Missing Models & 7,255 & 935,878,301 \\ \midrule
\textbf{Total} & \textbf{13,881} & \textbf{936,258,491} \\ \bottomrule
\end{tabular}
\label{tab:local_database_comparison}
% \vspace{-5mm}
\end{table}

\begin{table}[t!]
\centering
\caption{Task details and difficulty.}
\begin{tabular}{@{}lp{0.3\textwidth}p{0.12\textwidth}@{}}
\toprule
\textbf{ID} & \textbf{Task} & \textbf{Difficulty} \\ \midrule
T1 & Image Super-Resolution & Easy \\
T2 & Text-to-Image Generation & Moderate \\
T3 & Generate Image with Referenced Face & Advanced \\
T4 & Generate Image with Referenced Outline & Advanced \\
T5 & Virtual Try-On & Advanced \\ \bottomrule
\end{tabular}
\label{tab:task_details}
% \vspace{-5mm}
\end{table}

\section{User study}

\begin{table*}[t!]
\centering
\caption{Results from the user study. The first row shows the input images and text prompts (if applicable), and the second row presents the corresponding generated outputs.}
\label{tab:examples}
\resizebox{\textwidth}{!}{
\begin{tabular}{p{0.18\textwidth}|p{0.18\textwidth}|p{0.18\textwidth}|p{0.18\textwidth}|p{0.18\textwidth}}
\toprule
\textbf{T1: Image Super-Resolution} & \textbf{T2: Text-to-Image} & \textbf{T3: Generation with Referenced Face} & \textbf{T4: Generation with Referenced Outline} & \textbf{T5: Virtual Try-On} \\
\midrule
\includegraphics[width=\linewidth]{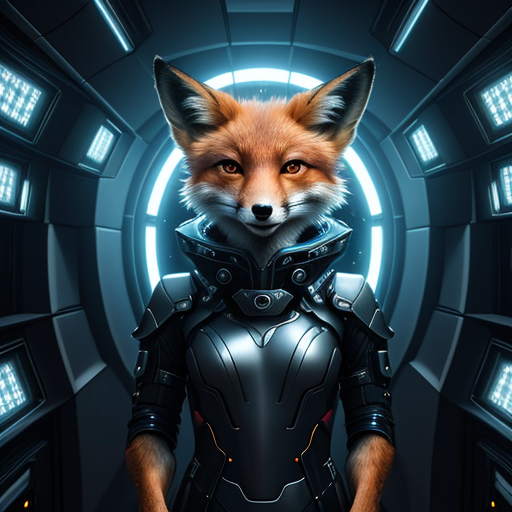} &
-- &
\includegraphics[width=\linewidth]{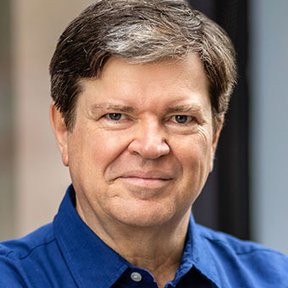} &
\includegraphics[width=\linewidth]{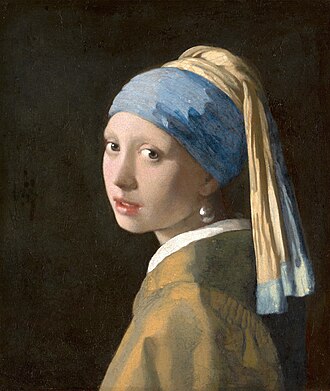} &
\includegraphics[width=0.48\linewidth]{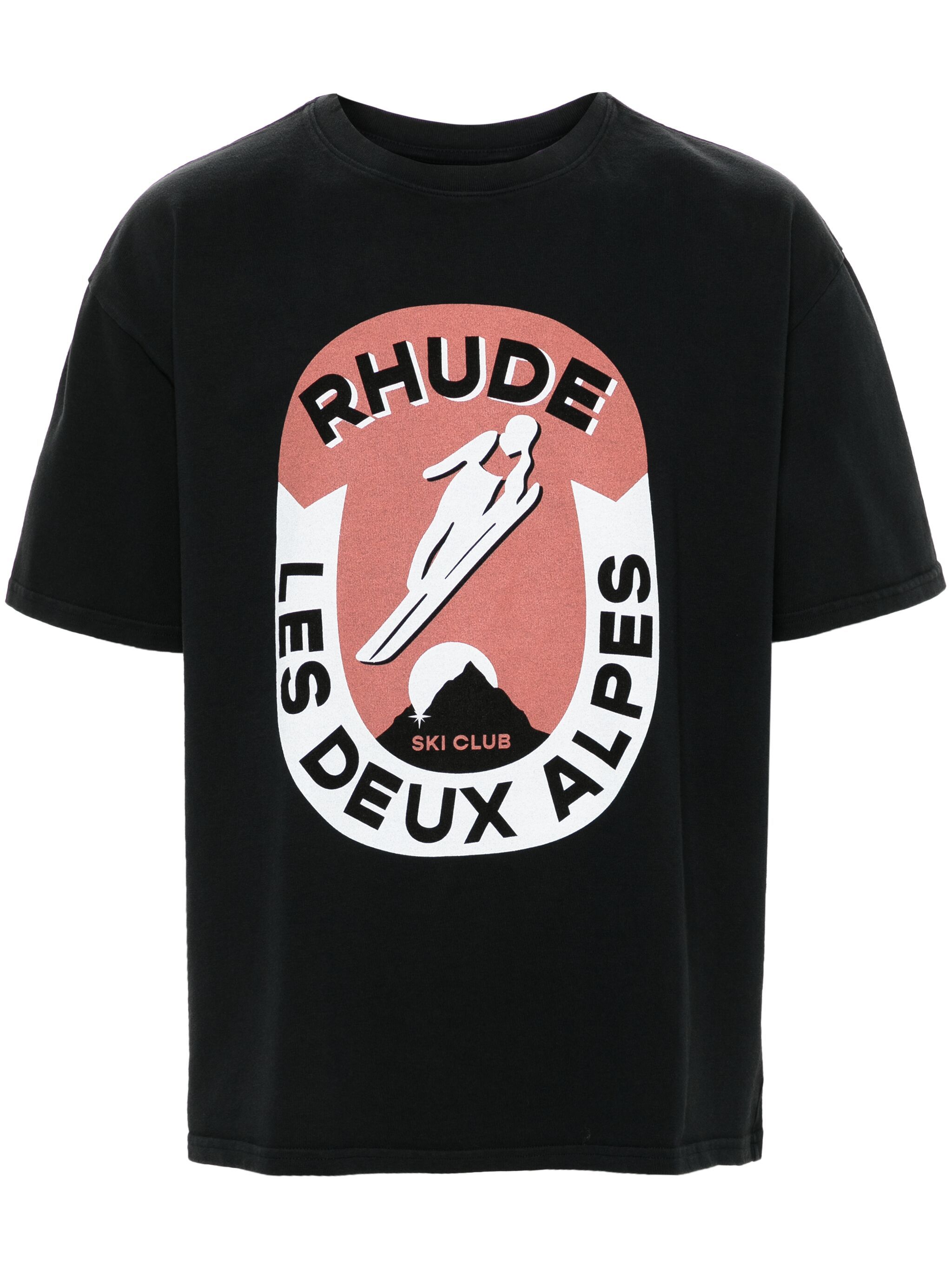}\hfill
\includegraphics[width=0.48\linewidth]{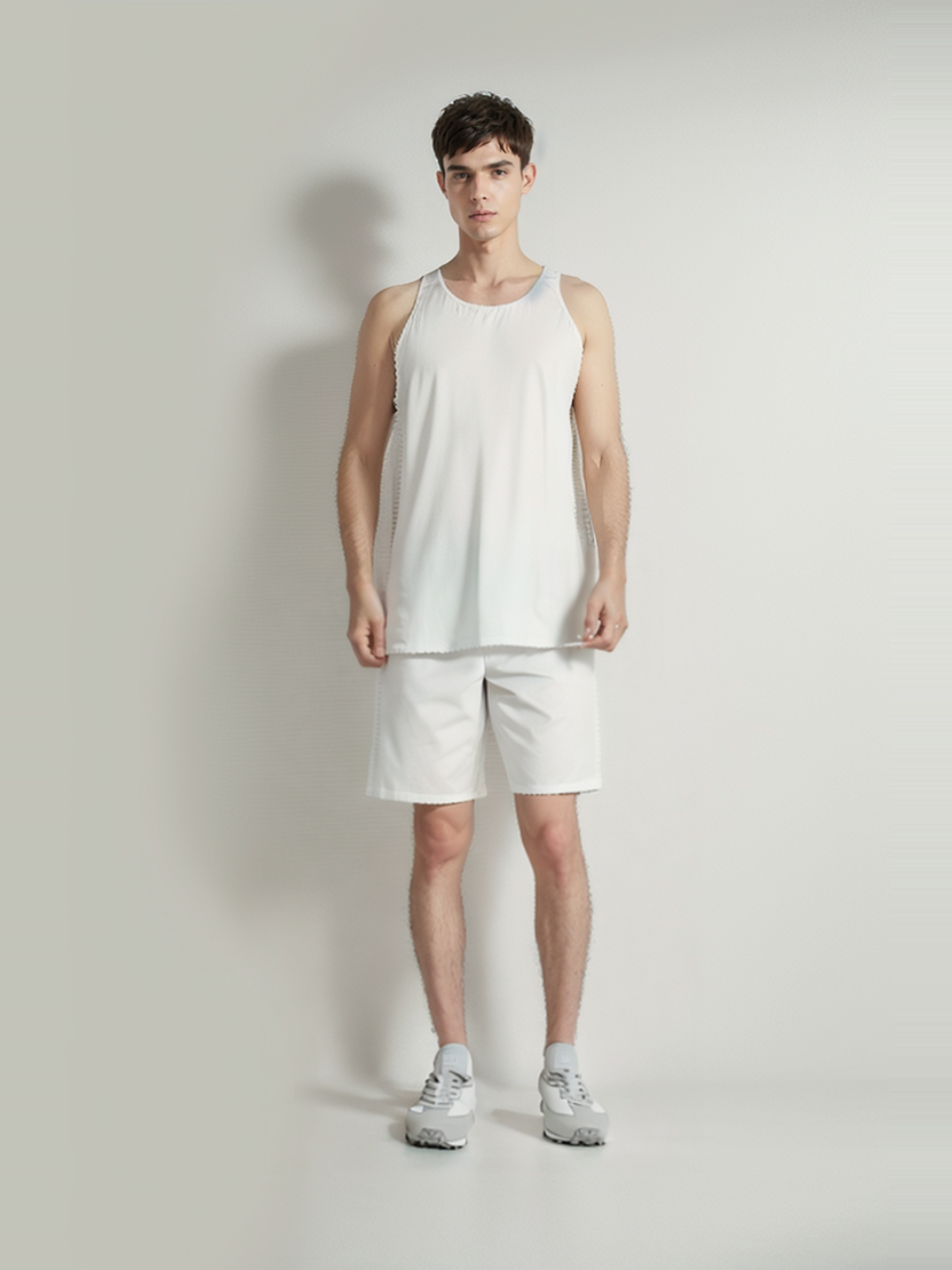} \\

-- &
a photo of a hyper sport car &
a photo of a man walking down a city sidewalk, wearing a blue shirt and beige pants, high resolution &
a photo of a female knight in medieval times, high resolution, highly detailed, 4k, cinematic, dramatic lighting &
-- \\
\midrule
\includegraphics[width=\linewidth]{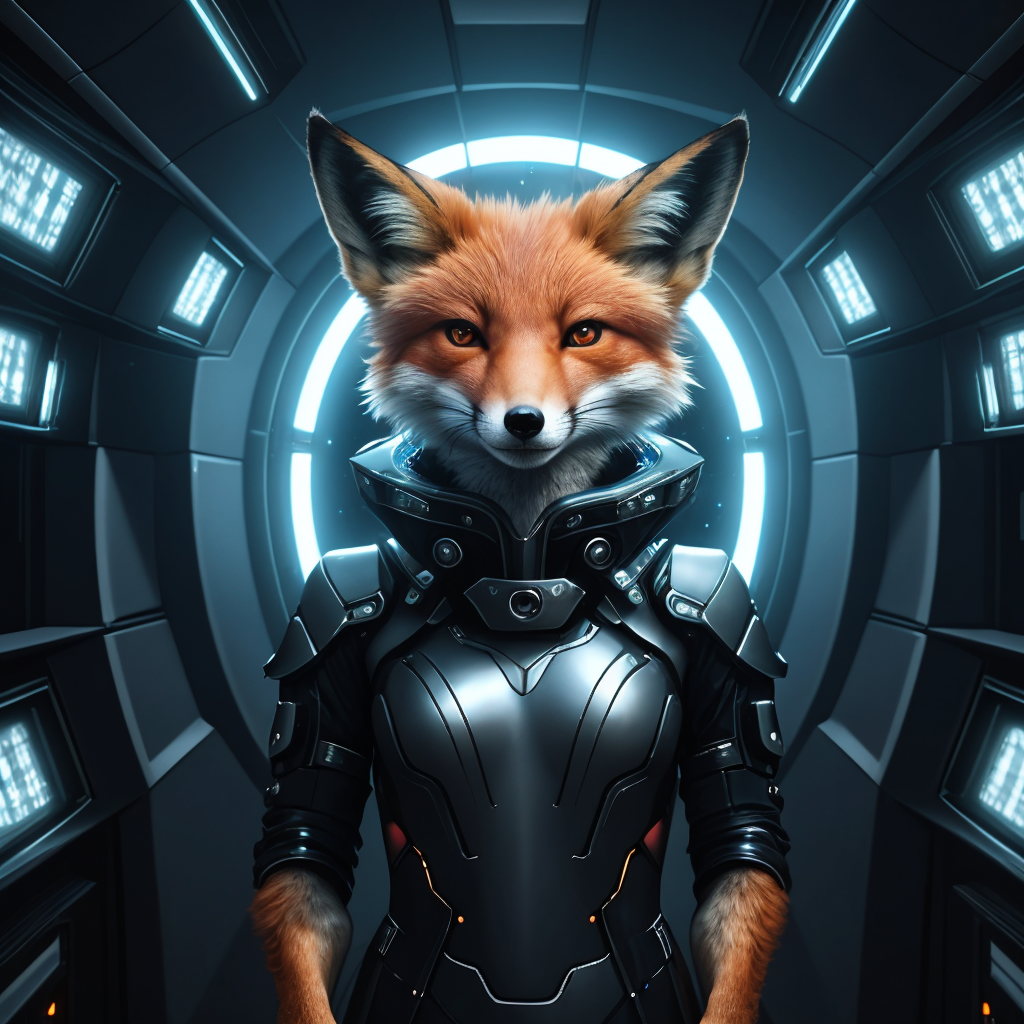} &
\includegraphics[width=\linewidth]{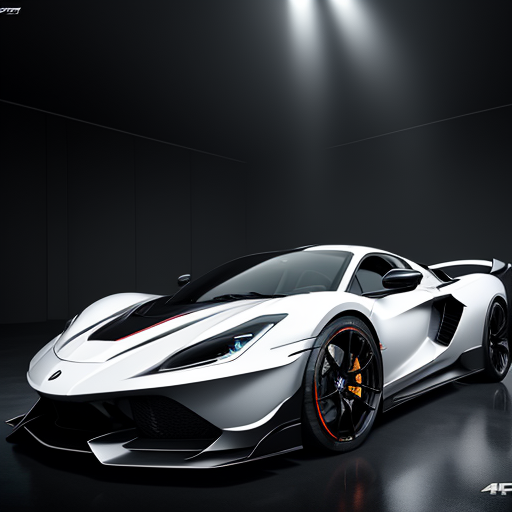} &
\includegraphics[width=\linewidth]{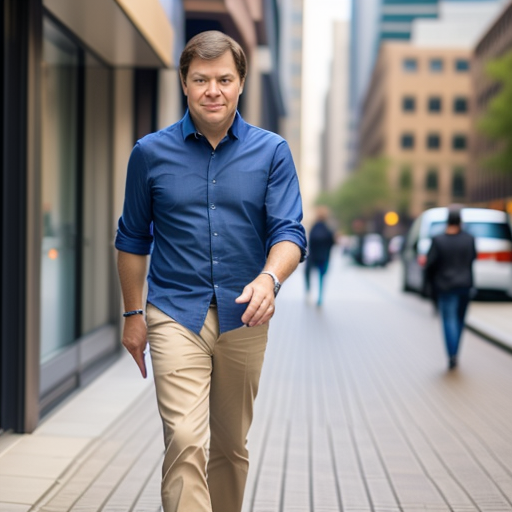} &
\includegraphics[width=\linewidth]{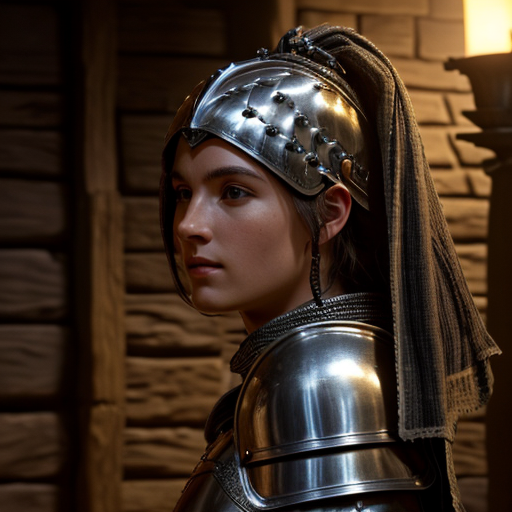} &
\includegraphics[width=\linewidth]{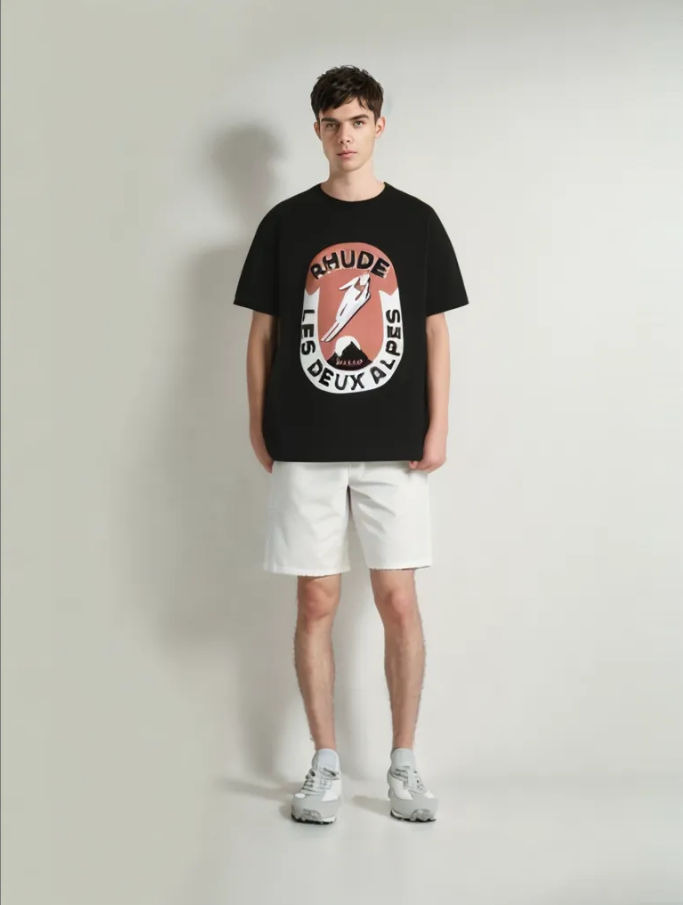} \\
\bottomrule
\end{tabular}
}
\end{table*}

We conducted a user study to evaluate GenFlow’s effectiveness in helping individuals with aphantasia visualize abstract concepts and generate personalized artistic outputs.

% We conducted a user study to gather  feedback on our proposed GenFlow. The framework aims to assist individuals with aphantasia in visualizing abstract concepts and producing personalized artistic outputs.

\subsection{Setup}

\subsubsection{Participants}

We invited 13 participants, denoted from P1 to P13, from a university, including 3 researchers and 10 STEM students, to experience with our system. All participants had programming skills ranging from intermediate to proficient. Among them, 38\% were experts in AI and computer vision, 38\% had intermediate knowledge, and 24\% were novices.

\subsubsection{Tasks}

Before the user study, we conducted a group interview to understand participants' typical tasks. One participant emphasized the time-consuming nature of tasks like image enhancement and generation, which often require multiple tools and manual adjustments. Another participant noted the challenges of generating realistic images from reference inputs, such as facial images or sketches, which demand expertise and fine-tuning. Based on these insights, we designed five tasks to streamline these processes, as shown in Table \ref{tab:task_details}.

\subsubsection{Apparatus and Procedure}

Each study session, conducted within our lab, involved participants performing assigned tasks on a provided laptop under our observation. Each session consisted of four sections within a 30-minute slot. After an introduction, participants followed step-by-step instructions in a video (10 minutes) and were free to ask questions for clarification. Next, participants completed the five study tasks detailed in Table \ref{tab:task_details}. Participants were encouraged to think aloud. At the end of each session, we collected feedback through surveys.

% \begin{table}[t!]
% \caption{Participants' self-reported roles and their expertise in programming and the computer vision task (rated 1-3), task completion time (in seconds)}
% \label{tab:participant_information}
% \begin{tabular}{@{}|c|ccc|ccccc|@{}}
% \toprule
% \textbf{ID} & \textbf{Role} & \textbf{Coding} & \textbf{AI} & \textbf{Task 1} & \textbf{Task 2} & \textbf{Task 3} & \textbf{Task 4} & \textbf{Task 5} \\ \midrule
% P1 & Researcher & 3 & 3 & 90 & 192 & 165 & 106 & 804 \\
% P2 & Researcher & 3 & 3 & 72 & 152 & 80 & 81 & 571 \\
% P3 & Student & 2 & 2 & 125 & 71 & 129 & 101 & 694 \\
% P4 & Student & 2 & 2 & 327 & 334 & 61 & 447 & 580 \\
% P5 & Student & 3 & 3 & 151 & 176 & 197 & 178 & 543 \\
% P6 & Student & 2 & 1 & 143 & 140 & 127 & 125 & 763 \\
% P7 & Student & 2 & 2 & 171 & 171 & 153 & 132 & 822 \\
% P8 & Student & 3 & 3 & 152 & 214 & 239 & 211 & 710 \\
% P9 & Student & 2 & 2 & 116 & 128 & 221 & 263 & 677 \\
% P10 & Student & 1 & 1 & 187 & 367 & 225 & 345 & 373 \\
% P11 & Researcher & 3 & 3 & 116 & 118 & 207 & 216 & 841 \\
% P12 & Student & 3 & 2 & 122 & 405 & 73 & 198 & 794 \\
% P13 & Student & 3 & 1 & 143 & 138 & 299 & 244 & 963 \\ \bottomrule
% \end{tabular}
% \end{table}

\subsection{Results}

\begin{table*}[t!]
\centering
\caption{Participant ratings of system aspects.}
\vspace{-3mm}
\begin{tabular}{@{}llc@{}}
\toprule
\textbf{Aspect}           & \textbf{Description}                                                                 & \textbf{Average Rating} \\ \midrule
Usefulness                & Effectiveness in reducing task completion time.                                       & 4.3 / 5                    \\ 
Friendly Interface        & Moderate user-friendliness with potential for improved intuitiveness.                 & 3.8 / 5                    \\ 
Ease of Use               & Straightforward and accessible navigation and usability.                              & 4.2 / 5                    \\ 
Performance               & Consistent and reliable processing speed and responsiveness.                         & 4.1 / 5                    \\ 
Overall Satisfaction      & General approval of functionality and performance.                                    & 4.0 / 5                    \\ \bottomrule
\end{tabular}
% \vspace{-5mm}
\label{tab:participant_ratings}
\end{table*}

\subsubsection{Task Completion}

All participants successfully completed all five tasks, with average completion times of 144.23, 205.07, 166.61, 193.31, and 662.08 seconds for Tasks 1 through 5, respectively.

Prompts were concise, typically under 25 words, and varied in style based on task complexity. For the easy task (T1: Image Super-Resolution), participants used direct commands like "increase resolution" [P13] or "enhance image quality" [P6]. The moderate task (T2: Text-to-Image Generation) encouraged more creative inputs, such as "generate an image of hypercar" [P5]. Advanced tasks required greater specificity. In T3 (Referenced Face Generation), prompts included "create an image with a face matching this reference" [P8]. In T4 (Referenced Outline Generation), examples were "generate image by filling this outline" [P12]. T5 (Virtual Try-On) involved precise instructions like "change clothes of the model’s figure" [P7].

Participants also used purpose-driven phrasing, e.g., "I want to create a portrait with this face in a medieval setting" [P1] and "I want to adjust the clothes for this person" [P2]. This variety in prompt styles reflected participants’ adaptability to different task difficulties. Simpler commands were sufficient for tasks like T1, while more complex and descriptive prompts were necessary for advanced tasks like T3, T4, and T5. This progression highlighted participants’ ability to adjust their prompting strategies based on task complexity.

\subsubsection{Insights}

Participants reflected on their experiences with the application, often comparing it to familiar tools like Photoshop, ChatGPT, and manual model pipelines. Their feedback revealed several critical observations:

\begin{itemize}
    \item \textbf{Efficiency Gains}: Participants, especially those with programming experience, praised the application's ability to streamline workflows. P1 remarked, “In a video file system, I haven’t seen anything that can do this,” emphasizing its unique functionality. P9 noted it “saves time” compared to self-hosted setups, while P13 described it as “faster” than manual alternatives. P4 highlighted the benefit of not needing to search for models, calling it a major time-saver.
    \item \textbf{User Interface}: The UI was commended for supporting rapid iteration. P2 observed that “iterations here are much faster with UI” due to the reduced need for complex prompts. The drag-and-drop interface stood out for ease of use, with P6 noting, “just drag and drop,” making it accessible for non-coders. P10 appreciated the system’s transparency and control, stating, “it shows the entire workflow and lets us configure it,” offering more control than generative AI tools.
\end{itemize}

% \begin{figure*}[t!]
%     \centering
%     \begin{subfigure}[t]{0.3\textwidth}
%         \centering
%         \includegraphics[width=\textwidth]{figures/web_search/bing_search.png}
%         \caption{Start Page}
%         \label{fig:bing_search}
%     \end{subfigure}
%     \hfill
%     \begin{subfigure}[t]{0.3\textwidth}
%         \centering
%         \includegraphics[width=\textwidth]{figures/web_search/result_search.png}
%         \caption{Result Search Page}
%         \label{fig:result_search}
%     \end{subfigure}
%     \hfill
%     \begin{subfigure}[t]{0.3\textwidth}
%         \centering
%         \includegraphics[width=\textwidth]{figures/web_search/huggingface_parse.png}
%         \caption{Hugging Face Download Page}
%         \label{fig:huggingface_parse}
%     \end{subfigure}
%     \caption{Download ThinkDiffusionXL Model Example by Web Exploration}
%     \label{fig:download_process}
% \end{figure*}

Participants broadly praised the tool’s automation, visual workflow design, and ease of customization. These features enabled users to reduce manual effort, transition seamlessly between tasks, and focus on creative or higher-level goals. The visual representation of workflows helped users, especially novices, better understand complex processes and experiment confidently. Support for multiple models and flexible workflow configurations enhanced the tool’s adaptability and long-term value. Examples created by participants in the user study are illustrated in Table \ref{tab:examples}.

% The user study highlighted several key strengths of the application, forming a robust foundation for its value proposition. Participants praised its workflow automation for enhancing efficiency by reducing manual effort and enabling seamless transitions between tasks, allowing users to focus on higher-order objectives. The visual representation of workflows promoted a deeper understanding of complex processes, fostering experimentation and innovation, particularly benefiting less experienced users by simplifying their learning curve. Its flexibility in customizing workflows to meet diverse needs, along with support for multiple models, was commended for ensuring long-term usability and adaptability. Compared to traditional tools, the application offered a faster, more intuitive experience with built-in AI capabilities, eliminating the need for external plugins and providing superior customization and dynamic workflow design. These features made it a versatile and efficient tool, appealing to both novice and advanced users.

% However, participants identified performance issues, such as slower processing speeds, which occasionally disrupted the user experience. Suggestions for improvement included clearer UI elements and faster performance to enhance usability further. By addressing these performance challenges, the application can solidify its position as an intuitive and efficient solution for users across all experience levels.

\subsubsection{Rating}

The system was evaluated through a survey of 13 participants, focusing on five key dimensions: Usefulness, Interface Friendliness, Ease of Use, Performance, and Overall Satisfaction. Average ratings were derived from participant responses. As shown in Table~\ref{tab:participant_ratings}, participants found the system effective in supporting task completion, particularly by reducing execution time and maintaining strong performance. While overall satisfaction and ease of use were high, feedback suggests that refining the user interface could further enhance usability and make the system more accessible to a wider range of users.

\subsubsection{Discussions}

The user study confirms GenFlow’s effectiveness in reducing task completion times and improving user satisfaction. Participants consistently praised its intuitive interface and streamlined workflows, reinforcing the value of our design approach. By utilizing a node-based architecture, GenFlow builds on familiar visual programming paradigms while offering the flexibility to accommodate rapid advances in AI technology. Additionally, Flow Pilot's intelligent retrieval system, grounded in recent progress in natural language processing and retrieval-augmented generation, positions GenFlow as a forward-looking tool in the AI art domain.

However, feedback from the participants also revealed key areas where the system could be improved to better support diverse user needs. Some users desired more precise workflow suggestions, indicating the need to enhance Flow Pilot’s natural language understanding and support multimodal queries (e.g., sketches or reference images). Others suggested that sharing and refining workflows collaboratively would support learning and innovation, highlighting the value of a community-driven repository. Additionally, varying comfort levels with node-based interfaces suggest that adaptive UI features, such as personalized layouts or usage-based recommendations, could improve usability for novices.

These findings point to the importance of making GenFlow more context-aware, collaborative, and adaptable, ensuring its continued effectiveness for a broad range of users as generative AI tools grow more complex.

% The future development will focus on several key areas. First, we will enhance Flow Pilot's natural language understanding capabilities to process more nuanced user queries, potentially incorporating multi-modal inputs such as sketches or reference images to refine workflow suggestions. Second, we plan to develop a community-driven workflow repository within GenFlow, enabling users to share, rate, and refine workflows. This collaborative ecosystem will accelerate innovation in generative art through collective expertise. Third, to further improve usability, we will explore adaptive interface features, including personalized node layouts based on user expertise levels. Finally, we will conduct larger-scale user studies to validate these enhancements across diverse demographic groups.
% These advancements will ensure GenFlow remains a versatile and user-centric platform, capable of meeting the needs of both novice creators and seasoned professionals in the rapidly evolving field of AI-driven art.

\section{Conclusion}
% This paper introduced GenFlow, a novel interactive modular framework that bridges the gap between user-friendliness and powerful customization in AI art generation \cite{sims2024creation}. We addressed the limitations of existing tools, which often lack fine-grained control or present a steep learning curve, by combining a node-based workflow editor with an intelligent workflow retrieval system. The integration of a chatbot interface and automated web search capabilities further simplifies the process of finding and applying relevant workflows. 

% Future development will focus on enhancing GenFlow's usability. This includes providing clearer error messages with actionable feedback, highlighting key inputs within workflows, and reducing visual clutter by default hiding intermediate nodes. These improvements will streamline workflows, particularly in complex projects, and make GenFlow more intuitive for all users.

This paper introduced GenFlow, a novel interactive modular framework that bridges the gap between user-friendliness and powerful customization in AI art generation. Our work takes a substantial step toward democratizing generative art by making advanced tools accessible to a broader audience without compromising the depth required by expert users.

Future work will focus on enhancing GenFlow’s workflow retrieval through improved language understanding and multimodal inputs, such as sketches and images, as well as developing a collaborative repository for sharing and refining workflows. These directions aim to foster community-driven innovation and adapt the system to diverse user expertise levels.

\section*{Acknowledgments}

% This research is supported by research funding from Faculty of Information Technology, University of Science, Vietnam National University - Ho Chi Minh City.

This research is funded by Vietnam National Foundation for Science and Technology Development (NAFOSTED) under Grant Number 102.05-2023.31.

\bibliographystyle{IEEEtran}
\balance
\bibliography{sample-base}

\end{document}